\documentclass{article}

\usepackage{microtype}
\usepackage{graphicx}
\usepackage{subfigure}
\usepackage{booktabs} % for professional tables

\usepackage{hyperref}

\usepackage[accepted]{icml2025}

\usepackage{amsmath}
\usepackage{amssymb}
\usepackage{mathtools}
\usepackage{amsthm}

\usepackage{algorithm}
\usepackage{algorithmic}

\usepackage{multicol}
\usepackage{multirow}
\usepackage{pifont}
\usepackage{newfloat}
\usepackage{listings}
\floatstyle{ruled}
\newfloat{listing}{tb}{lst}{}
\floatname{listing}{Listing}

\usepackage[capitalize,noabbrev]{cleveref}

\theoremstyle{plain}

\theoremstyle{definition}

\theoremstyle{remark}

\usepackage[textsize=tiny]{todonotes}

\icmltitlerunning{Learning to Compress Contexts for \\ Efficient Knowledge-based Visual Question Answering}

\begin{document}

\twocolumn[
\icmltitle{Learning to Compress Contexts for \\ Efficient Knowledge-based Visual Question Answering}

% It is OKAY to include author information, even for blind
% submissions: the style file will automatically remove it for you
% unless you've provided the [accepted] option to the icml2025
% package.

% List of affiliations: The first argument should be a (short)
% identifier you will use later to specify author affiliations
% Academic affiliations should list Department, University, City, Region, Country
% Industry affiliations should list Company, City, Region, Country

% You can specify symbols, otherwise they are numbered in order.
% Ideally, you should not use this facility. Affiliations will be numbered
% in order of appearance and this is the preferred way.
\icmlsetsymbol{equal}{*}

% \author {
%     % Authors
%     Weixi Weng\textsuperscript{\rm 1},
%     Jieming Zhu\textsuperscript{\rm 2},
%     Hao Zhang\textsuperscript{\rm 2},
%     Xiaojun Meng\textsuperscript{\rm 2},
%     Rui Zhang\textsuperscript{\rm 3},
%     Chun Yuan\textsuperscript{\rm 1},
%     }
% \affiliations {
%     % Affiliations
%     \textsuperscript{\rm 1}Tsinghua Shenzhen International Graduate School\\
%     \textsuperscript{\rm 2}Huawei Noah's Ark Lab\\
%     \textsuperscript{\rm 3}ruizhang.info\\
%     wengwx22@mails.tsinghua.edu.cn, \{jamie.zhu, zhang.hao3, xiaojun.meng\}@huawei.com, rayteam@yeah.net, yuanc@sz.tsinghua.edu.cn
% }

\begin{icmlauthorlist}
\icmlauthor{Weixi Weng}{yyy}
\icmlauthor{Jieming Zhu}{comp}
\icmlauthor{Xiaojun Meng}{comp}
\icmlauthor{Hao Zhang}{comp}
\icmlauthor{Rui Zhang}{sch}
\icmlauthor{Chun Yuan}{yyy}
%\icmlauthor{}{sch}
%\icmlauthor{}{sch}
\end{icmlauthorlist}

\icmlaffiliation{yyy}{Tsinghua Shenzhen International Graduate School}
\icmlaffiliation{comp}{Huawei Noah's Ark Lab}
\icmlaffiliation{sch}{School of Computer Science \& Tech, Huazhong University of Science and Technology}

\icmlcorrespondingauthor{Chun Yuan}{yuanc@sz.tsinghua.edu.cn}

% You may provide any keywords that you
% find helpful for describing your paper; these are used to populate
% the "keywords" metadata in the PDF but will not be shown in the document
\icmlkeywords{Machine Learning, ICML}

\vskip 0.3in
]

% this must go after the closing bracket ] following \twocolumn[ ...

% This command actually creates the footnote in the first column
% listing the affiliations and the copyright notice.
% The command takes one argument, which is text to display at the start of the footnote.
% The \icmlEqualContribution command is standard text for equal contribution.
% Remove it (just {}) if you do not need this facility.

%\printAffiliationsAndNotice{}  % leave blank if no need to mention equal contribution
% \printAffiliationsAndNotice{\icmlEqualContribution} % otherwise use the standard text.

\begin{abstract}
Multimodal large language models (MLLMs) have demonstrated great performance on visual question answering (VQA). 
When it comes to knowledge-based Visual Question Answering (KB-VQA), MLLMs may lack the specialized domain knowledge needed to answer questions, necessitating the retrieval of necessary information from external knowledge sources.
Previous works like Retrival-Augmented VQA-v2 (RAVQA-v2) focus on utilizing as much input information, such as image-based textual descriptions and retrieved knowledge, as possible to improve performance, but they all overlook the issue that with the number of input tokens increasing, inference efficiency significantly decreases, which contradicts the demands of practical applications.
To address this issue, we propose \textbf{R}etrieval-\textbf{A}ugmented MLLMs with \textbf{C}ompressed \textbf{C}ontexts (RACC). 
RACC learns to compress and aggregate retrieved knowledge for a given image-question pair, generating a compact modulation in the form of Key-Value (KV) cache to adapt the downstream frozen MLLM, thereby achieving effective and efficient inference.
RACC achieves a state-of-the-art (SOTA) performance of 63.92\% on OK-VQA. Moreover, it significantly reduces inference latency by 22.0\%-59.7\% compared to the prominent RAVQA-v2. 
Abundant experiments show RACC's broad applicability. It is compatible with various off-the-shelf MLLMs and can also handle different knowledge sources including textual and multimodal documents.
\end{abstract}

\section{Introduction}

Multimodal large language models (MLLMs) have attracted wide research attention, demonstrating great zero-shot performances among various visual question answering (VQA) datasets. 
However, in practical applications, generating accurate answers to specific questions necessitates not just a precise grasp of image content, but also human commonsense or domain-specific knowledge. This category of VQA tasks is known as knowledge-based VQA (KB-VQA). Given that knowledge parameterized within MLLMs is static and limited, utilizing an external knowledge source to furnish necessary information to MLLMs emerges as a dependable strategy for addressing KB-VQA challenges.

In previous studies of KB-VQA, a line of works \cite{hu2023promptcap, khademi2023mm, an2024knowledge} utilize knowledge from very large MLLMs (GPT-4) or LLMs (ChatGPT, GPT-3) to aid in answering question.
However, the static knowledge in such models may become out-of-date, and the models may generate incorrect content due to hallucinations, particularly in specific domains. What's more, the high cost associated with using these models is also a notable downside.
Another line of research which retrieves knowledge from updateable external knowledge sources, such as knowledge graphs \cite{speer2017conceptnet}, documents \cite{luo2021weakly}, \emph{etc.}, is more reliable and better aligns with the needs of real-world applications. 

Retrieval Augmented VQA-v2 (RAVQA-v2) tackles the problem of KB-VQA by performing straightforward retrieval-augmented generation (RAG) on MLLMs.
However, it has a notable shortcoming, \emph{i.e.}
% 1. Incorrect supervised signals during finetuning. 
% The $K$ retrieved documents from knowledge sources are likely to include irrelevant ones, which cannot provide the necessary knowledge for the MLLM to answer the question, leading to incorrect answers. RAVQA-v2 framework even forces the MLLM to generate correct answers based on irrelevant documents, which imposes incorrect supervised signals on the MLLM and may cause it to generate information that is absent from the documents after finetuning.
low efficiency during inference. In the inference process of RAVQA-v2, the $K$ retrieved documents are first concatenated with the image-question pair and inputted into the MLLM to obtain $K$ candidate answers, then the final answer is selected from the $K$ candidates by their joint probabilities. The process is undoubtedly very time-consuming and resource-intensive. Moreover, the retrieved documents can be quite long and often contain a lot of redundant information, which can further exacerbate the problem of low inference efficiency.

Inference efficiency is a key concern in practical applications of MLLMs. However, previous work, including RAVQA-v2, has focused on how to use as much knowledge as possible to improve the accuracy of answers while neglecting the fact that inference efficiency significantly declines as the number of input tokens increases.

Therefore, we aim to design an innovative RAG framework based on MLLMs, which can utilize the information of retrieved knowledge in an effective and efficient manner to improve MLLMs' inference efficiency for KB-VQA.

Furthermore, RAVQA-v2 and many previous works \cite{luo2021weakly, lin2022retrieval, lin2022revive} on KB-VQA primarily focus on using textual documents as external knowledge sources, and there has been relatively less research on using multimodal documents as knowledge sources. 
However, multimodal documents are an important knowledge resource in real-world applications, which can provide knowledge in handling KB-VQA \cite{raffel2020exploring, hu2023reveal}. What's more, MLLMs inherently have the ability to directly comprehend multimodal knowledge. Therefore, we aim to explore the effects of multimodal documents in RAG applications based on MLLMs. 

% Our proposed framework can effectively utilize the inherent capabilities of MLLMs to employ both textual and multimodal documents as knowledge sources in a unified, effective, and efficient manner, thereby enhancing the KB-VQA performance of MLLMs.
% Therefore, we craft a multimodal knowledge source from the Wikipedia Image-Text dataset and refer to it as the WIT knowledge source hereafter. We fine-tune MLLMs on OK-VQA \cite{marino2019ok} based on RAVQA-v2 using this knowledge source, and the experimental results are shown in Table \ref{tab: ravqa reproduction}. The results demonstrate that WIT can also provide the necessary information for various MLLMs to answer image-based questions in OK-VQA. Multimodal documents are quite a common type of knowledge source in real-world applications, and We believe that exploring such kind of knowledge sources is beneficial for RAG research on MLLMs.
In this paper, we propose RACC, \emph{i.e.} \textbf{R}etrieval-\textbf{A}ugmented MLLMs with \textbf{C}ompressed \textbf{C}ontexts, an effective and efficient RAG framework for KB-VQA based on MLLMs. RACC consists of three phases, namely compression learning, information aggregation, and modulation generation.
% Our proposed framework first leverages a frozen hyperMLLM to learn to compress retrieved documents into short soft prompts. Then, we design an elaborate aggregator module to aggregate compressed prompts. Finally, a set of Multi-Layer Perceptrons (MLPs) is used to generate a compact modulation in the form of Key-Value (KV) cache to adapt the downstream frozen baseMLLM. With the compact modulation, the baseMLLM can utilize the information in the retrieved documents in a highly efficient manner. 
Our contributions can be summarized as follows:
\begin{itemize}
    \item RACC is the first work to integrate prompt compression technology with KB-VQA, proposing an innovative framework for efficient RAG on MLLMs. We identify four key issues in training RACC and propose four corresponding methods to address them.
    \item RACC achieves excellent performance comparable to many competitive baselines on two KB-VQA datasets at a very low cost, reaching a state-of-the-art (SOTA) performance of 63.92\% on the OK-VQA dataset.
    \item RACC achieves outstanding enhancements in terms of time and space efficiency. For time efficiency, RACC can save 22.0-59.7\% of inference latency compared to RAVQA-v2. For space efficiency, RACC supports pre-saving documents that occupy a large storage footprint in the form of compressed prompts to save disk space.
    \item RACC can be applied to various off-the-shelf MLLMs, but also can handle different knowledge sources such as textual documents and multimodal documents.
\end{itemize}

\begin{figure*}[bt]
    \centering
    \includegraphics[width=1\textwidth]{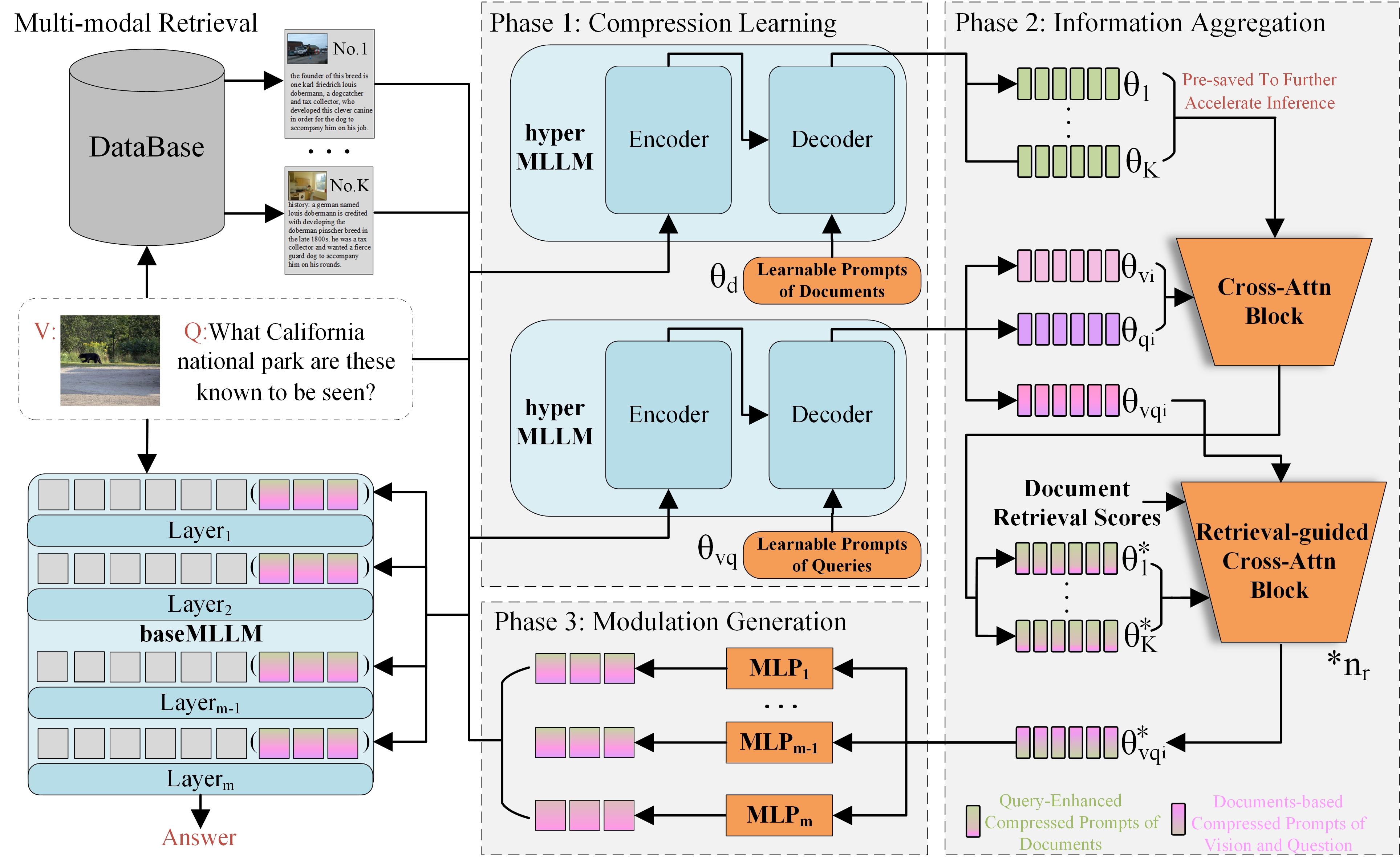}
    \vspace{-2em}
    \caption{
    The structural framework of RACC. An image and a question are first input into the multimodal retriever to retrieve $K$ relevant documents. During the compression learning phase, the $K$ documents, image, and question are input into HyperMLLM to obtain their corresponding compressed prompts. In the information aggregation phase, the obtained compressed prompts are aggregated to form the document-based compressed prompts of vision and question. In the modulation generation phase, the output of the aggregation network is processed by a set of MLPs to obtain the KV cache for each layer of the downstream BaseMLLM. At the same time, the BaseMLLM receives the image and question as input and generates the final answer.}
    % is outlined as follows: Initially, an image-question pair is input into the frozen multimodal retriever, which retrieves $K$ documents from the knowledge source. Subsequently, the hyperMLLM compresses both the retrieved knowledge and the image-question pair. The aggregator then aggregates these compressed prompts, while a set of multi-layer perceptrons (MLPs) generates a P-Tuning v2 modulation based on the aggregated compressed prompts. This generated modulation adapts the downstream frozen baseMLLM to conduct effective and efficient KB-VQA. }
    \label{fig: framework structure}
    \vspace{-0.5em}
\end{figure*}

\section{Related Work}
\subsection{Multimodal Large Language Models}
Multimodal large language models (MLLMs) bridge the gap between text and other modalities and unify understanding of different modalities. Although recent MLLMs \cite{alayrac2022flamingo, li2023blip, li2023instructblip, bai2023qwen, liu2023visual, hu2024minicpm} have demonstrated excellent zero-shot performance in VQA tasks, they require the assistance of external knowledge sources for questions that require specialized domain knowledge.

In this work, we investigate two main architectures of MLLMs: encoder-decoder MLLMs including BLIP2-FlanT5XL \cite{li2023blip} and InstructBLIP-FlanT5XL \cite{li2023instructblip}, and the decoder-only MLLMs including InstructBLIP-Vicuna7B and miniCPM-v2 \cite{hu2024minicpm}.

\subsection{Knowledge-based Visual Question Answering}
Knowledge-based Visual Question Answering (KB-VQA), anchored by the given images, works to answer questions that require external knowledge. KB-VQA is an important multimodal task and has garnered widespread attention.

Early KB-VQA works \cite{luo2021weakly, marino2021krisp} train specialized VQA models with designated knowledge sources such as ConceptNet KnowledgeGraph \cite{speer2017conceptnet}, textual document sources like Google Search \cite{luo2021weakly} or WikiData \cite{vrandevcic2014wikidata}, and image sources like Google Image Search \emph{etc.}. However, these methods often demonstrate limited performance.

With the emergence of pretrained large language models (LLMs), they have become a focus of research in this field. A series of works utilize very large LLMs like GPT-3 \cite{brown2020language}, chatGPT to generate auxiliary knowledge or directly answer the questions \cite{gui2021kat, lin2022revive, yang2022empirical, shao2023prompting, hu2023promptcap}, achieving performance breakthroughs. 
Due to the high costs associated with using very large LLMs like GPT-3, and the fact that the knowledge within them can be outdated and incorrect, another line of work focuses on how to conduct retrieval-augmented generation (RAG) on smaller LLMs with various external knowledge bases for KB-VQA \cite{gui2021kat, gao2022trig, lin2022retrieval, lin2024fine, hu2023reveal}. However, the above methods all face the same problem: they require converting images into textual descriptions such as captions, object tags, \emph{etc.,} so that LLMs can understand \cite{gui2021kat, gao2022trig, shao2023prompting, lin2024fine}, which may result in loss of critical visual information, but also significantly increases the number of input tokens, leading to a notable increase in inference latency.

With the advent of multimodal large language models (MLLMs), the aforementioned problems have been perfectly solved. Recent works have made great progress by utilizing MLLMs. A line of works proposes to combine MLLMs and LLMs together \cite{khademi2023mm, xenos2023simple, an2024knowledge, liang2024endtoendvideoquestionanswering, liang2025reasvqa}. MM-Reasoner \cite{khademi2023mm} leverages vision APIs and rationales generated by GPT-4 to fine-tune MLLMs such as Flamingo. 
% ReasVQA \cite{liang2025reasvqa} leverages even imperfect reasoning processes generated from additional MLLMs to guide VideoQA models. 
RAVQA-v2 \cite{lin2024fine} is the first work to build a simple RAG framework on top of MLLMs. However, it suffers from low efficiency during the inference stage and doesn't study the impact of multimodal documents on RAG applications of MLLMs.

\subsection{Prompt Compression}
Given the inherent redundancy in natural language, prompt compression methods have been extensively studied to improve the efficiency of LLM inference. Prompt compression can be categorized into task-aware and task-agnostic methods. Since the generation of compressed prompts that perform well across diverse tasks is particularly challenging, we focus on the task-aware prompt compression paradigm. 

An important line of work \citet{jiang2023llmlingua, pan2024llmlingua, jiang2023longllmlingua} estimates the importance of the tokens within the original prompts by the information-based metrics and removes redundant tokens. \citet{xu2024recomp} trains a compressor model, which cuts redundant tokens in the passage based on the question. 

In addition to the above methods, which detect and remove inherent redundant tokens in long contexts at the natural language level, a series of works aims at compressing long contexts into parameters, leveraging the capability of LLMs to implicitly eliminate redundant information within long contexts.
\citet{mu2024learning} supposes that each prompt is composed of a task instruction part and a content part, and finetunes LLMs to compress the task instruction part into several gist tokens. \citet{chevalier2023adapting, wang2024greater, tack2024online} proposed to compress long contexts into compact summary vectors, parameters of a Lora-module and KV Cache, respectively. Such kind of methods have already demonstrated significant research value and strong research potential. In this paper, we extend these methods to the KB-VQA task.

% As MLLMs have seen widespread adoption, to the best of our knowledge, research on prompt compression for MLLM remains an unexplored field.

\begin{table*}[h]\fontsize{9pt}{10pt}\selectfont
    \centering
    \begin{tabular}{c | c | c | c | c }
         \hline
         Model & Image-base Textual Description & Base Model & knowledge source & VQA Accuracy \\
         \hline
         \multicolumn{5}{l}{\textit{Specialized baselines}} \\
         \hline
         KRISP &  &  & C & 38.35 \\         
         VRR & Caption &  & GS & 45.08 \\
         MALI &  &  & miniGPT4 + C & 56.69 \\ 
         REVIVE & Caption + Object Tags &  & WD + GPT-3 & 58.00 \\
         REVEAL &  & T5-Large & WIT + CC + WD + V2 & 59.10 \\
         \hline
         \multicolumn{5}{l}{\textit{Baselines on LLMs}} \\
         \hline
         KAT & Caption + Object Tags & T5-large & W & 44.25 \\
         KGenVQA & Caption & UnifiedQA & PNP & 45.40 \\ 
         PICa & Caption + Object Tags & GPT-3 & & 48.00 \\
         RA-VQA & OCR + Caption + Object Tags &  T5-large & GS & 51.22 \\
         KAT-Ensemble & Caption + Object Tags & T5-large & W + GPT-3 & 54.41 \\
         RA-VQAv2 & OCR + Caption + Object Tags &  T5-large & GS & 54.85 \\
         Prophet & Caption &  GPT-3 & MCAN & 58.27 \\
         PromptCap & Caption &  GPT-3 & ICE (16) & 60.40 \\
         \hline
         \multicolumn{5}{l}{\textit{Baselines based on MLLMs}} \\
         \hline
         PaLI &  &  PaLI-15B & & 56.50 \\
         Flamingo &  &  Flamingo & & 57.80 \\
         BLIP2 &  & BLIP2-FlanT5XL & & 31.76 \\
         % 17 & BLIP2 (finetuned) & BLIP2 (T5-xl) & OCR + Caption + Object Tags & 55.44 \\
         % RA-VQAv2 &  & BLIP2 (FlanT5XL) & WIT & 56.44 \\  
         RA-VQAv2 &  & BLIP2-FlanT5XL & GS & 60.40 \\
         RA-VQAv2 &  & InstructBLIP-FlanT5XL & GS & 62.90 \\
         % 22 & InstructBLIP (finetuned) &  & InstructBLIP (T5-xl) & & 57.32 \\
         % 23 & RA-VQAv2 (zero-shot) &  & InstructBLIP (T5-xl) & WIT & 33.80\\
         % 11 & KCR &  OCR + Caption + Object Tags & BLIP2 (T5-xl) + T5-xl & GS + MCAN & 65.10 \\
         \hline
         \multicolumn{5}{l}{\textit{Baselines based on both LLMs and MLLMs}} \\
         \hline
         MM-Reasoner & OCR +  Caption + Object Tags... & Flamingo & GPT-4 & 59.20 \\
         ASB & Caption & LLAMA-2 & PNP + ICE (14)  & 59.07 \\
         DKA & Caption & LLAMA-2 & PNP + ChatGPT + ICE (14) & 62.10 \\
         \hline 
         \multicolumn{5}{l}{\textit{Our proposed framework based on MLLMs}} \\
         \hline
         RACC-\texttt{homo}  &  & BLIP2-FlanT5XL & WIT & 55.07 \\
         RACC-\texttt{homo} &  & InstructBLIP-FlanT5XL & WIT & 59.17 \\
         RACC-\texttt{homo} &  & BLIP2-FlanT5XL & GS & 55.26 \\
         RACC-\texttt{homo} &  & InstructBLIP-FlanT5XL & GS & 59.49 \\ 
         RACC-\texttt{hetero} &  & BLIP2-Vicuna7B & GS & 61.65 \\
         RACC-\texttt{hetero} &  & InstructBLIP-Vicuna7B & GS &  \textbf{63.92} \\
         \hline
    \end{tabular}
        \caption{Model Performance on the OK-VQA dataset. Knowledge source abbreviations: C: ConceptNet; CC: CC12M; V2: VQA-2; W: Wikipedia; WD: WikiData; WIT: Wikipedia Image-Text; GS: Google Search; GI: Google Images; ICE: In-context Examples; PNP: Plug-and-Play VQA captioner \cite{tiong2022plug}. 
        Please refer to Section \ref{sec: modulation generation} for the implications for RACC-\texttt{homo} and RACC-\texttt{hetero}.
        % In RACC-\texttt{homo}, the hyperMLLM and baseMLLM share the same structures and weights, while in RACC-\texttt{hetero}, they differ in either structure or weights. 
        In the last two rows of results of the RACC-\texttt{hetero}, the hyperMLLM used is InstructBLIP-FlanT5XL. }
    \label{tab: okvqa results}
    \vspace{-1em}
\end{table*}

\section{Methods}
In this section, we introduce our proposed RACC, \emph{i.e.} \textbf{R}etrieval-\textbf{A}ugmented MLLM with \textbf{C}ompress \textbf{C}ontexts. 
RACC can be divided into three phases: compression learning, information aggregation, and modulation generation, which will be detailed in the corresponding subsections. 
Based on a profound understanding of the task and an analysis of the shortcomings in RAVQA-v2, we identify four key issues that need to be addressed in training RACC and further propose four methods to address them, which is elaborated in Subsection \ref{sec: compression learning} and \ref{sec: information aggregation}.
\subsection{Problem Setup}\label{sec: problem setup}
A typical VQA dataset can be divided into three components: images, questions, and answers, which can be represented using the notation $\{v, q, a\}_n$.
Following \citet{lin2024fine}, we consider a realistic scenario of KB-VQA: an MLLM that takes an image $v_i$ and its related question $q_i$ as input, where the knowledge required to answer the question is supplied by external knowledge sources. 
In this paper, we study two important knowledge sources in real-world applications: multimodal documents and textual documents.

We utilize an off-the-shelf frozen multimodal retriever to retrieve $K$ documents from the given knowledge source conditioned on the provided image and question. The $K$ retrieved documents is denoted as $\{d_i\}^K = \{d^1_i, d^2_i, \ldots, d^K_i\}$.

The confidence scores that the multimodal retriever outputs for each of the $K$ documents, \emph{i.e.} the document retrieval scores, are denoted as $\{p_i\}^K = \{p^1_i, p^2_i, \ldots, p^K_i\}$. The MLLM needs to leverage these retrieved documents to provide the correct answer to the question $q_i$ based on $v_i$.
\subsection{Phase 1: Compression Learning}\label{sec: compression learning}
The first phase of RACC is to compress the retrieved document into soft prompts of a specified length. \citet{phang2023hypertuning} and \citet{tack2024online} introduce the idea of amortized-based meta-learning into the online learning task, which brings us inspiration. We utilize MLLMs of encoder-decoder architecture (such as BLIP2-FlanT5XL \cite{li2023blip}, InstructBLIP-FlanT5XL \cite{li2023instructblip}, \emph{etc.}) with a set of learnable prompts to compress the input information.

We denote the hyperMLLM by $\textsc{M}_{hyper}$ and represent the encoder and decoder within the hyperMLLM by $\textsc{Enc}_{hyper}$ and $\textsc{Dec}_{hyper}$, respectively. Each retrieved document is first fed into $\textsc{Enc}_{hyper}$. Subsequently, $\textsc{Dec}_{hyper}$ takes the output from the encoder and the learnable prompts $\theta_d$ as inputs, and the resulting output is the compressed prompts corresponding to that document. After such a
compression process, the contextual information of each document is preserved in its compressed prompts. Consequently, in subsequent processing steps, the number of tokens that need to be processed for each document is reduced from the original token count of the document to the predefined length of $\theta_d$. The compressing process can be denoted as follows:
% \begin{equation}
\begin{equation}
    \theta^k_i = \textsc{M}_{hyper}(d^k_i, \theta_d) = \textsc{Dec}_{hyper}(\textsc{Enc}_{hyper}(d^k_i), \theta_d)
\end{equation}
% \end{equation}
$\theta^k_i$ denotes the compressed prompts of the document $d^k_i$. 

We also compress the given image-question pair in a similar manner.
% The key differences are that we first process the image and question independently using the hyperMLLM, and subsequently, we concatenate them for an additional round of compression. The reason behind this design choice will be elaborated in the following section. 
Note that the predefined learnable prompts for compressing the image-question pairs are different from $\theta_{d}$, which we denote as $\theta_{vq}$, the compression process of the given image-question pair is denoted as follows:
\begin{equation}
    \begin{split}
        % &\theta_{v_i} = \textsc{M}_{hyper}(v_i, \theta_{vq}), \\
        % &\theta_{q_i} = \textsc{M}_{hyper}(q_i, \theta_{vq}), \\
        &\theta_{vq_i} = \textsc{M}_{hyper}(\textsc{concat}(v_i, q_i), \theta_{vq})
    \end{split}
\end{equation}

We name this phase the compression learning phase, as the learnable prompts need to be trained to acquire the ability to compress document context information. We then identify two key issues in this phase and propose corresponding solutions, as described below:

\textbf{Issue 1: How to initialize the learnable prompts?}

\underline{\textit{Analysis:}}
First, the lengths of the two sets of learnable prompts, \emph{i.e.} $\theta_{d}$ and $\theta_{vq}$, are critical in the compression learning process. If the prompts are too short, the amount of compressed semantic information that they can retain will be insufficient. In contrast, if the prompts are too long, the difficulty of training increases significantly.

We suggest determining the lengths of two sets of learnable prompts by experiments. To determine the optimal lengths of these prompts, we conducted a series of comparative experiments. We treat the lengths of $\theta_{d}$ and $\theta_{vq}$ as hyperparameters and systematically explore their impact on the compression learning process.

Additionally, the initialization weights of the learnable prompts also play a crucial role in the hyperMLLM's ability to compress inputs, especially in the early training stage.

\underline{\textit{Solution:}}
Therefore, we propose a method to initialize learnable prompts called \textbf{P}rompt \textbf{I}nitialization with hard \textbf{P}rompt \textbf{E}mbeddings (PIPE). 
We begin by manually designing two sets of hard prompts. For example, the hard prompt corresponding to $\theta_d$ is ``Summarize the key information of the given passage in a concise manner." The hard prompts are then processed by the tokenizer and the embedding layer of hyperMLLM to generate their embeddings. Subsequently, these embeddings are used to initialize the weights of the learnable prompts $\theta_d$ and $\theta_{vq}$. 

\textbf{Issue 2: How to deal with the irrelevant documents?}

\underline{\textit{Analysis:}}
The documents retrieved by the multimodal retriever may sometimes be completely irrelevant to the given image and question. Even if they are relevant, they may not provide the model with useful information to give the correct answer. 

However, RAVQA-v2 \cite{lin2024fine} does not consider the impact of irrelevant documents and treats both irrelevant documents and useful documents in the same way, \emph{i.e.} concatenating retrieved documents with corresponding image-question pairs and inputting them into MLLMs to calculate the loss based on the correct answer. 
In clearer terms, RAVQA-v2 forces MLLMs to generate correct answers even based on irrelevant documents, which imposes incorrect supervised signals on the MLLM and can significantly harm the training process.

\underline{\textit{Solution:}}
Therefore, to avoid the negative impact on the compression learning process caused by the involvement of irrelevant documents in training, we propose a method called \textbf{P}seudo-\textbf{R}elevance-based \textbf{B}ackpropagation \textbf{D}ropout (PRDB), which is introduced as follows:

Following \citet{luo2021weakly, lin2024fine}, we consider a document to be pseudo-relevant if it contains any of the human-annotated answers. 
After all $K$ retrieved documents are converted into compressed prompts, we apply a stop gradient operation, \emph{i.e.} \textsc{stopgrad}($\textsc{M}_{hyper}({d^k_i}, \theta_d)$) to the compressed prompts of those documents which are considered pseudo-irrelevant. Only the gradients of the compressed prompts of the pseudo-relevant documents will be backpropagated to the learnable prompts.
In this way, after loss calculation and gradient backpropagation, PRDB prevents the gradients of the compressed prompts corresponding to irrelevant documents from leading the learnable prompts' weights to update in the wrong direction, ensuring the stability of the compression learning phase.

\subsection{Phase 2: Information Aggregation}\label{sec: information aggregation}
After the compression learning phase, we now have a set of compressed prompts of the retrieved documents $\{ \theta_i \}^K$, $\theta_{v_i}$, $\theta_{q_i}$ and $\theta_{vq_i}$. In this section, we detail the process of using $\theta_{vq_i}$ to aggregate useful semantic information from $\{ \theta_i \}^K$. We further identify two key issues in designing the information aggregation process and propose two corresponding methods to address them. We name this phase as the information aggregation phase.

\textbf{Issue 3: How to enhance the semantic information in the compressed prompts of documents associated with the given image and question? }\label{DCSE}

\underline{\textit{Analysis:}}
The retrieved documents often contain a large number of redundant tokens, with only a small portion of tokens being relevant to the question and the image, which can be useful for answering the question and should be paid more attention to. Even after compression, the compressed prompts can still contain redundant semantic information. We think that leveraging the information from the image and the question to enhance the semantic information in the compressed prompts of documents is essential. 
% We first explore the relationship between V (images) and Q (questions) from the task and model perspectives.

% From the perspective of KB-VQA, both images and questions are crucial components and are complementary to each other. However, when it comes to a specific image-question pair, the importance of the image and question may differ. In some cases, it is necessary to retrieve relevant documents based on key details present in the image but not mentioned in the question. Conversely, when the question contains substantial information, it is more appropriate to focus on retrieving documents closely related to the question. Given that the retrieved documents contain a lot of tokens and the portion of content relevant to the image or question might be relatively small, we believe that enhancing the semantic information in retrieved documents using image and question-based information is essential.

% From the perspective of MLLMs, the relationship between images and questions is unequal, which is reflected in the number of tokens they occupy. In nearly all MLLMs, image features are converted into fixed-length tokens, such as 32, while the number of tokens for questions is often smaller. This disparity may result in the learnable prompts focusing too much on the image part in the early training stage. 

\underline{\textit{Solution:}}
Based on the above, we propose a method called DCSE, \emph{i.e.} \textbf{D}ecoupled \textbf{C}ompression for \textbf{S}emantics \textbf{E}nhancement. 
Our method begins by decoupling the given image-question pair $(v_i, q_i)$ and input them separately into the hyperMLLM with $\theta_{vq}$, generating $\theta_{v_i}$ and $\theta_{q_i}$, which can be denoted as follows:
\begin{equation}
    \begin{split}
        &\theta_{v_i} = \textsc{M}_{hyper}(v_i, \theta_{vq}) \\
        &\theta_{q_i} = \textsc{M}_{hyper}(q_i, \theta_{vq})
    \end{split}
\end{equation}

Compared with $\theta_{vq_i}$ obtained by jointly compressing the image-question pair, $\theta_{v_i}$ and $\theta_{q_i}$ derived from decoupled compression can better preserve the semantic information contained in both the image and the question. In the retrieved documents, the semantic information related to either the image or the question is of significance for answering the question. Therefore, we employ the concatenated $\theta_{v_i}$ and $\theta_{q_i}$ to enhance the semantic information of $\{\theta_i\}^K$ via cross-attention. Here, we adopt a original cross-attention block, where $\{\theta_i\}^K$ serves as the query, and the concatenated $\theta_{v_i}$ and $\theta_{q_i}$ act as both the key and the value. 
% This method also helps prevent learnable prompts from overly focusing on image tokens during the process of learning coupled and decoupled compression of image-question pairs. 
Denoting a original cross-attention block \cite{vaswani2017attention} as \textbf{CA}, the computational process of DCSE is denoted as follows:
\begin{equation}
    \{ \theta ^*_i\}^K =\textbf{CA}( \{ \theta_i\}^K, \textsc{concat}(\theta_{v_i}, \theta_{q_i}))
\end{equation}
$\{ \theta^*_i\}^K$ represents the query-enhanced compressed prompts of documents.

\textbf{Issue 4: How to utilize the document retrieval scores to guide the aggregation process?}

\underline{\textit{Analysis:}}
In the retrieval process, most retrievers assign a confidence score to each retrieved document based on metrics such as embedding similarity. In the following text, we use "document retrieval score" to represent this score.

While judging whether a document can truly answer a question based on an image is difficult, the retrieval score offers a reliable metric for this purpose. Based on the principles of retrieval mechanisms, documents with higher retrieval scores are generally considered to provide more relevant and useful information for the given image and question. \citet{lin2024fine} utilizes the retrieval scores as a reference metric for selecting the final answer in the inference process, but it does not use this crucial metric in the training process. 

\underline{\textit{Solution:}}
Based on the above, we propose a method called \textbf{R}etrieval-\textbf{G}uided \textbf{C}ross-\textbf{A}ttention (RGCA).
RGCA is an improvement of the original cross-attention mechanism, designed to gather the semantic information from $\{ \theta ^*_i\}^K$ that can assist in answering questions into $\theta_{vq_{i}}$, guided by $\{p_i\}^K$. 
% In RGCA, $\theta_{vq_{i}}$ functions as the query and $\{ \theta ^*_i\}^K$ serves as the key and value.
RGCA not only considers the embedding similarity between compressed prompts but also assigns more attention to the compressed prompts of documents with higher retrieval scores. Refer to the appendix for the pseudo code of RGCA. We denote a retrieval-guided cross-attention block as $\textbf{CA}_r$, whose forward pass is denoted as:
\begin{equation}
    \theta^*_{vq_i} =\textbf{CA}_r(\theta_{vq_i}, \{\theta_i^*\}^K,\{p_i\}^K)
\end{equation}
The number of retrieval-guided cross-attention blocks contained in our framework is set to $n_r=3$.

\subsection{Phase 3: Modulation Generation}\label{sec: modulation generation}
After the information aggregation phase, we have the documents-based compressed prompts of vision and question, which is denoted as $\theta^*_{vq_i}$. Then we convert $\theta^*_{vq_i}$ into a P-Tuning v2 modulation for the downstream baseMLLM, which involves adding a small amount of KV cache at each layer of the baseMLLM. Given that different layers of MLLM process information at varying levels of abstraction and complexity, we employ a set of $m$ Multi-Layer Perceptrons (MLPs) for projecting $\theta^*_{vq_i}$ into the additional KV cache of each layer in the baseMLLM, where $m$ is the number of layers within the baseMLLM. Here, we denote the generated P-Tuning v2 modulation as $\Theta_i$ and the frozen baseMLLM as $\textsc{M}_{base}$. 

Our framework can be optimized in an end-to-end manner using the loss function $\mathcal{L}$, namely the language modeling loss based on the ground truth answer:
\begin{equation}
    \underset{\theta_d, \theta_{vq}, h}{\min} \frac{1}{N}\sum^{N}_{i=1}\mathcal{L}(\textsc{M}_{base}(v_i,q_i; \Theta_i), a_i)
\end{equation}
$h$ includes a single \textbf{CA} block, $n_r$ $\textbf{CA}_r$ blocks and a set of MLPs. $N$ is the training batch size. RACC offerss two variants: RACC-\texttt{homogeneous} and RACC-\texttt{heterogeneous}, abbreviated as RACC-\texttt{homo} and RACC-\texttt{hetero}, respectively.
In the setup of RACC-\texttt{homo}, the hyperMLLM and the baseMLLM are identical, which means that the MLLM learns to compress contexts for itself. For RACC-\texttt{hetero}, the hyperMLLM and the baseMLLM differ in either structure or weight.

\section{Experiments}\label{sec: efficiency}
\subsection{Datasets and Knowledge Sources}
We evaluate our framework on OK-VQA \cite{marino2019ok}, which is the most widely studied KB-VQA dataset. 
% The average number of words in the questions of OK-VQA is 16.14. The average number of words in the passages of Google Search is 59.76.
We also conduct experiments on AOK-VQA \cite{schwenk2022okvqa}, which is the successor of OK-VQA.

In terms of the knowledge source, following \citet{lin2024fine}, we adopt Google Search \cite{luo2021weakly} for OK-VQA and AOK-VQA, which is a textual document base comprised of nearly 200 thousand documents. We also carefully curated a multimodal document source from the Wikipedia Image-Text dataset \cite{srinivasan2021wit} for OK-VQA to show that our framework also works well with multimodal knowledge sources. We use GS and WIT to refer to these two knowledge sources.

We use FLMR \cite{lin2024fine} and PREFLMR \cite{lin2024preflmr} as retrievers for document retrieval. The FLMR retriever was used to retrieve information from the GS knowledge source, while the PREFLMR retriever was used to retrieve information from the WIT knowledge source.

\subsection{Training Setup}

Most of the experiments are conducted on a 32G V100 GPU. The chosen optimizer is AdamW. During the first 1000 steps of training, the learning rate linearly increases from $10^{-5}$ to $10^{-4}$. Subsequently, a cosine-decaying scheduler is applied to the learning rate to reduce it from $10^{-4}$ to 0. The batch size is set to 2. The hyperparameter $K$, \emph{i.e.} the number of retrieved documents for each image-question pair, is always set to 5. Note that during the training process of RACC, all parameters of the hyperMLLM, baseMLLM, and multimodal retrievers are kept frozen.

\subsection{Evaluation}
We evaluate the performance of our framework using the official VQA Accuracy \cite{marino2019ok}. Let $a_i$ be the list of human-annotated answers of the given image-question pair $(v_i, q_i)$, and $y_i$ be the model's outputs. The VQA accuracy for $(v_i, q_i)$ is calculated as follows:
\begin{equation}
    \textsc{VQAAccuracy}(a_i, y_i) = min(\frac{\#S(y_i)}{3}, 1)
\end{equation}
where $\#S(y_i)$ is the occurrence of $y_i$ in $a_i$. The VQA accuracy on the entire dataset is obtained by averaging the accuracy of all image-question pairs.

\begin{table}[t]
    \centering
    \begin{tabular}{c | c | c c}
        \hline
        \multirow{2}{*}{Method} & \multirow{2}{*}{Base Model} & \multicolumn{2}{c}{Direct Answer} \\  
        \multirow{2}{*}{} & & Val & Test \\
        \hline
        ClipCap & & 30.9 & 25.9 \\
        LXMERT & & 30.7 & 25.9  \\
        KRISP & &  33.7 & 27.1  \\
        KGenVQA & UnifiedQA &  39.1 & - \\
        GPV-2 & T5-Large &  48.6 & 40.7 \\
        REVEAL & T5-Large &  52.2 & - \\
        PromptCap & GPT-3 &  56.3 & 59.6 \\
        MM-Reasoner & Flamingo + i-Code &  - & \textbf{60.2} \\ 
        ASB & LLAMA-2 &  58.6 & 57.5 \\
        RACC-\texttt{homo} & InstructBLIP-FlanT5XL &  \textbf{62.1} & 58.1 \\
        \hline
    \end{tabular}
    \caption{The results on the AOK-VQA dataset. We use the GS knowledge for AOK-VQA here. }
    \vspace{-0.5em}
    \label{tab: aokvqa results}
\end{table}

\subsection{Comparative Study}
In this section, we will elaborate on the advantages of RACC compared to previous works from the following three aspects: performance, cost, and inference efficiency. 

First of all, RACC outperforms many competitive baselines. The performance comparison of RACC with other competitive baselines on the OK-VQA dataset is presented in Table \ref{tab: okvqa results}. Based on InstructBLIP-FlanT5XL, RACC-\texttt{homo} with GS as the knowledge source reaches an accuracy of 59.65\%. With WIT as the knowledge source, our framework achieves 59.17\%. When adopting RACC-\texttt{hetero}, with InstructBLIP-FlanT5XL as the hyperMLLM and InstructBLIP-Vicuna7B as the baseMLLM, we achieve a state-of-the-art (SOTA) accuracy of 63.92\%.

The results of AOK-VQA are shown in Table \ref{tab: aokvqa results}. Since the GS knowledge source we use for AOK-VQA is not designed for it, the documents in GS may not provide the required knowledge for all questions in AOK-VQA. However, RACC-\texttt{homo} based on InstructBLIP-FlanT5XL still achieves a state-of-the-art (SOTA) accuracy of 62.1\% on the validation set. The performance on the test set is 58.1\%.

\begin{table}[b]
    \centering
    % \vspace{-1em}
    \begin{tabular}{c | c | c c}
        \hline
        {} & \multirow{2}{*}{RAVQA-v2} & \multicolumn{2}{c}{RACC-\texttt{homo}} \\  
        {} & \multirow{2}{*}{} & w/o pre & w pre \\
        \hline
        % Training Time Per Sample (s) & 0.800 &  1.002 \\
        
        Eval Time (s) & 1.1242 & 0.8768 & 0.4576 \\

        Disk Usage (M) & 6.9680 & 6.9680 & 0.6280 \\
        % VQA Accuracy & 58.77 & 59.17 & 59.17 \\

        % Memory Storage Occupied & \\
        \hline
    \end{tabular}
    \caption{Comparison of inference efficiency between RAVQA-v2 and RACC when adopting the WIT knowledge source. ``Eval time" and ``Disk Usage" are measured for a single image-question pair input. ``w pre" indicates pre-saving the compressed prompts of retrieved documents before inference. The MLLM used in both two frameworks is InstructBLIP-FlanT5XL.}
    \label{tab: comparison of cost}
\end{table}

In terms of cost, our work has notable advantages. First, we do not utilize any image-based textual descriptions provided by external APIs or models \cite{gui2021kat, lin2022retrieval, an2024knowledge}, such as captions, object tags, OCR \emph{etc.} Second, RACC does not use any very large LLMs (ChatGPT, GPT-3) or MLLMs (GPT-4) but still achieves excellent performance even with small-scale MLLMs.

The inference efficiency is the main concern of this paper.
RACC demonstrates significant advantages in inference efficiency compared to RAVQA-v2, which is shown in Table \ref{tab: comparison of cost}.
In Table \ref{tab: comparison of cost}, the inference efficiency and disk usage of RACC are demonstrated under two scenarios: "w/o pre" and "w pre". Here, "w pre" refers to the scenario where the compressed encodings of retrieved documents are pre-saved before the inference process, which is shown in Figure \ref{fig: framework structure}. We also present the inference efficiency and disk usage of RAVQA-v2. When pre-saving compressed prompts, we achieve a substantial reduction of 59.7\% in inference latency and 91.0\% in disk space usage compared to RAVQA-v2. Even without pre-saved compressed prompts, the inference latency can still be reduced by 22.0\%.

\subsection{Ablation Studies}\label{sec: ablations studies}
We propose four methods to improve the aggregation process of compressed contexts and conduct ablation studies to verify their effectiveness. The settings and results of ablation studies are depicted in Table \ref{tab: ablation studies} and Table \ref{tab: length of prompts}. Note that we adopt RACC-\texttt{homo} with the GS knowledge source in the ablation studies, where the hyperMLLM and baseMLLM are both initialized from InstructBLIP-FlanT5XL. Refer to the appendix for more details of the ablation studies.

Firstly, comparing lines 2 and 3, as well as lines 7 and 8, we can observe that the PIPE method brings improvements of 0.31\% and 0.42\% under different settings. From the difference between lines 3 and 4 in Table \ref{tab: ablation studies}, we observe that the DCSE method brings an improvement of 0.77\%. On the other hand, the RGCA method results in a performance gain of 0.37\%, as shown in lines 3 and 5. Last but not least, the performance difference between lines 6 and 8 shows that the PRDB method leads to a performance gain of 0.54\%.

We also investigate into the settings of some important hyperparameters.
We first explore how to set the length of learnable prompts (\emph{i.e. $L(\theta_{vq})$ and $L(\theta_d)$}), and the results are shown in Table \ref{tab: length of prompts}. We select the best configuration, setting $L(\theta_{vq})$ and $L(\theta_d)$ to 12 and 16, respectively. All other experiments in this paper are conducted using this configuration. In the appendix, we provide comparative experiments on the hyperparameter $K$.

\begin{table}[bt]\fontsize{9pt}{10pt}\selectfont
    \centering
    \begin{tabular}{c c c c c c}
        \hline
        No. & {PIPE} & {DCSE}& {RGCA} & {PRDB} & {VQA Accuracy (\%)} \\  
        \hline
        1 &  &  &  &  & 57.60 \\
        2 &  &  &  & \ding{51} & 58.18 (+0.58) \\
        3 & \ding{51} &  & & \ding{51} & 58.49 (+0.89) \\
        4 & \ding{51} & \ding{51} & & \ding{51} & 59.26 (+1.66)\\
        5 & \ding{51} & & \ding{51} & \ding{51} & 58.86 (+1.26) \\
        6 & \ding{51} &\ding{51} & \ding{51} &  & 58.95 (+1.35) \\
        7 & &\ding{51} & \ding{51} & \ding{51} & 59.07 (+1.47)\\
        8 & \ding{51} &\ding{51} & \ding{51} & \ding{51} & \textbf{59.49 (+1.89)} \\
        \hline
    \end{tabular}
    \caption{The results of ablation studies of RACC. The GS knowledge source is adopted here. The ablation experiments are conducted based on RACC-\texttt{homo} with InstructBLIP-FlanT5XL. }
    \label{tab: ablation studies}
\end{table}

\begin{table}[t]
    \centering
    \begin{tabular}{c c c}
        \hline
        {$L(\theta_{vq})$} & {$L(\theta_d)$} & {VQA Accuracy} \\ 
        \hline
        8 & 12 & 58.77 \\
        8 & 16 &  58.83\\
        12 & 12 & 58.96  \\
        12 & 16 & \textbf{59.07} \\
        12 & 20 & 58.56\\
        % {w PIPE} & 12 & 16 & 59.49 \\
        \hline
    \end{tabular}
    \caption{RACC-\texttt{homo}'s results of the comparative experiments on the length of the predefined learnable prompts $\theta_{vq}$ and $\theta_{d}$. The two sets of learnable prompts are randomly initialized here.
    }
    \label{tab: length of prompts}
\end{table}

\subsection{Broad Applicability of RACC}
RACC shows broad applicability from multiple aspects. 

1. RACC can utilize different types of knowledge sources to aid its efficient RAG process. We evaluate RACC with two knowledge sources, \emph{i.e.} WIT and GS, which represent multimodal documents and textual documents.

2. RACC can leverage any off-the-shelf multimodal retriever for retrieval, and our proposed RGCA method enables RACC to benefit from advancements in multimodal retrieval technology.

3. RACC can be applied to any off-the-shelf MLLMs. We further conduct experiments under the setup of RACC-\texttt{hetero} and present the results in Table \ref{tab: different structure}. Experiments show that RACC-\texttt{hetero} performs well across different baseMLLMs. The setup of RACC-\texttt{hetero} is also of practical significance: 
When it is not feasible to directly fine-tune the baseMLLM due to resource constraints, our framework can still work by adopting a much smaller hyperMLLM to adapt the larger frozen baseMLLM. 

\begin{table}[h]
    \centering
    \begin{tabular}{ c | c }
        \hline
        baseMLLM & VQA Accuracy \\  
        \hline
        miniCPM-v2 & 48.21 \\
        BLIP2-FlanT5XL & 54.91 \\
        InstructBLIP-FlanT5XL & 59.49 \\
        BLIP2-Vicuna7B &  61.65\\
        InstructBLIP-Vicuna7B &  \textbf{63.92}\\
        \hline
        % WIT & InstructBLIP & BLIP2 & 55.08 \\
        % WIT & BLIP2 & BLIP2 & 54.91 \\
    \end{tabular}
    \caption{RACC-\texttt{hetero}'s experimental results on OK-VQA using different MLLMs as the baseMLLM. The hyperMLLM is fixed as InstructBLIP-FlanT5XL here.}
    \label{tab: different structure}
\end{table}

\section{Conclusion}
In this paper, we propose \textbf{R}etrieval-\textbf{A}ugmented MLLMs with \textbf{C}ompressed \textbf{C}ontexts (RACC). RACC has achieved the following accomplishments in the area of KB-VQA:

1. RACC achieves competitive performance at a very low cost on challenging KB-VQA datasets. 

2. As the first work to explore how to conduct efficient RAG on MLLMs for KB-VQA tasks, RACC provides a reliable way that not only reduces inference latency but also significantly saves disk space.

3. RACC is applicable to different MLLMs and various kinds of external knowledge sources. 

With the rapid development of RAG technology and MLLMs, we believe that inference latency is a key concern in practical applications, which has often been overlooked in previous KB-VQA works. We hope our research will provide some inspiration for future work in this field.

% In the unusual situation where you want a paper to appear in the
% references without citing it in the main text, use \nocite
\nocite{langley00}

\bibliography{example_paper}
\bibliographystyle{icml2025}

%%%%%%%%%%%%%%%%%%%%%%%%%%%%%%%%%%%%%%%%%%%%%%%%%%%%%%%%%%%%%%%%%%%%%%%%%%%%%%%
%%%%%%%%%%%%%%%%%%%%%%%%%%%%%%%%%%%%%%%%%%%%%%%%%%%%%%%%%%%%%%%%%%%%%%%%%%%%%%%
% APPENDIX
%%%%%%%%%%%%%%%%%%%%%%%%%%%%%%%%%%%%%%%%%%%%%%%%%%%%%%%%%%%%%%%%%%%%%%%%%%%%%%%
%%%%%%%%%%%%%%%%%%%%%%%%%%%%%%%%%%%%%%%%%%%%%%%%%%%%%%%%%%%%%%%%%%%%%%%%%%%%%%%
\newpage
\appendix
\onecolumn
\section{RAVQA-v2 reproduction}
Since RAVQA-v2 does not investigate the effects of multimodal documents in its retrieval-agumented generation process, we replicate RAVQA-v2 using our crafted WIT knowledge source based on two types of MLLMs. 

The experimental results are shown in Table \ref{tab: ravqa reproduction}. Specifically, we follow all the experiment setup mentioned by \citet{lin2024fine}, such as the prompt template "Question: \{\} Knowledge: \{\} Answer:", where the Knowledge section includes both text tokens and image tokens transformed by Qformer. It is evident that the WIT knowledge source can, to some extent, provide the necessary knowledge for the OK-VQA dataset.

Additionally, we can observe that when using WIT as the knowledge source, the replication results of RAVQA-v2 across different MLLMs indicate that omitting the image information leads to better fine-tuning performance.

We believe this may be due to the following two reasons: (1) The inclusion of images directly increases the number of tokens in the knowledge section, making it more challenging for the MLLM to identify key information from the provided knowledge. (2) The use of images likely introduces redundant information that is unrelated to the document content, as images themselves contain a wealth of information. Furthermore, incorporating images also increases the time required for both training and inference based on RAVQA-v2. 

Overall, although RAVQA-v2 theoretically supports multimodal document-based knowledge sources, experiments show that its performance with multimodal documents is not satisfactory.

\begin{table}[h]
    \centering
    \begin{tabular}{c c c}
        \hline
        {MLLM} & {Knowledge Source} & {VQA Accuracy (\%)} \\  
        \hline
        \multirow{3}{*}{BLIP2} & {No Knowledge} & 54.10 \\
        \multirow{3}{*}{} & {WIT} & 56.26 \\
        \multirow{3}{*}{} & {WIT (Text Only)} & 56.44 \\
        % \multirow{4}{*}{} & {*Google Search (Text Only)} &  60.37 \\
        \hline
        \multirow{3}{*}{InstructBLIP} & {No Knowledge} & 57.32 \\
        \multirow{3}{*}{} & {WIT} & 58.77 \\
        \multirow{3}{*}{} & {WIT (w/o Image)} & 59.12 \\
        % \multirow{4}{*}{} & {Google Search (Text Only)} &  65.43 \\
        \hline
    \end{tabular}
    \caption{The performance of two types of MLLMs after finetuned on OK-VQA with the WIT knowledge source, based on the RAVQA-v2 framework.}
    \label{tab: ravqa reproduction}
\end{table}

\section{Complete Ablation Studies}
We identify four key issues in training RACC and further propose four methods to solve the issues. Extensive ablation studies are conducted to verify the effectiveness the proposed four methods. 

We conduct experiments on two types of MLLMs, \emph{i.e.} BLIP2-FlanT5XL and InstructBLIP-FlanT5XL. Due to page constraints of the paper content, we only present the ablation results for InstructBLIP-FlanT5XL in Table \ref{tab: ablation studies}. We adopt RACC-\texttt{homo} in the experiments of ablation studies, where the hyperMLLM and baseMLLM are both initialized from the models listed in the ”MLLM” column of Table \ref{tab: ablation studies}.

For RACC-\texttt{homo} based on BLIP2-FlanT5XL, comparing
lines 2 and 3, as well as lines 7 and 8, we can observe that the PIPE method brings improvements of 0.28\% and 0.44\% under different settings. The PIPE method is also effective for the InstructBLIP-based RACC-\texttt{homo}, leading to performance gains of 0.31\% and 0.42\% as shown by lines 10 and 11, as well as lines 15 and 16.

From the differences between lines 3 and 4, as well as lines 11 and 12 in Table 5, we can observe the effectiveness
of the DCSE method, bringing improvements of 0.55\% and 0.77\% respectively. The RGCA method, on the other hand, results in improvements of 0.28\% and 0.37\%, as shown in lines 3 and 5, as well as lines 11 and 13 in Table 5. 

Last but not least, the performance differences between
lines 6 and 8 and lines 14 and 16 in Table 5 show that the PRDB method leads to performance gains of 0.39\% and 0.54\% based on the two types of MLLMs. 

\begin{figure*}[htb]
    \centering
    \includegraphics[width=0.5\textwidth]{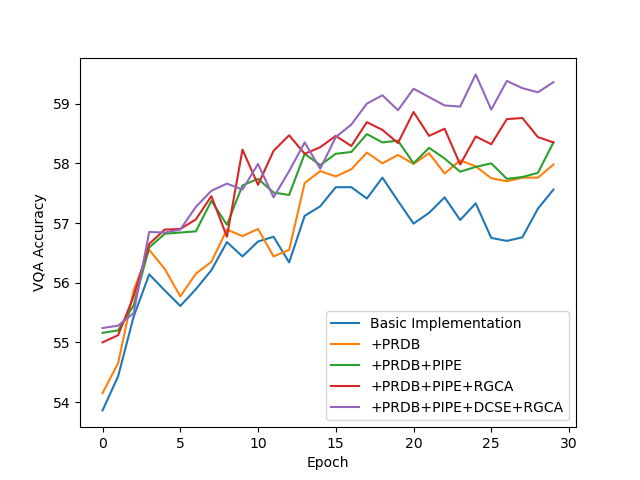}
    \caption{The VQA accuracy on the validation set during the training process of the framework using different methods.}
    \label{fig: curves}
\end{figure*}

We also plot the VQA accuracy on the validation set during the training process of the framework using different methods, as shown in Figure \ref{fig: curves}.

\begin{table*}[t]
    \centering
    \begin{tabular}{c c c c c c c}
        \hline
        {MLLM} & No. & {PIPE} & {DCSE}& {RGCA} & {PRDB} & {VQA Accuracy (\%)} \\  
        \hline
        \multirow{7}{*}{BLIP2-FlanT5XL} & 1 &  &  &  &  & 53.77 \\
        \multirow{7}{*}{} & 2 &  &  &  & \ding{51} & 54.20(+0.43) \\
        \multirow{7}{*}{} & 3 & \ding{51} &  & & \ding{51} &  54.48 (+0.71)\\
        \multirow{7}{*}{} & 4 & \ding{51} & \ding{51} & & \ding{51} & 55.03 (+1.26) \\
        \multirow{7}{*}{} & 5 & \ding{51} & & \ding{51} & \ding{51} &  54.76 (+0.99)\\
        \multirow{7}{*}{} & 6 & \ding{51} &\ding{51} & \ding{51} &  & 54.81 (+1.04)\\
        \multirow{7}{*}{} & 7 &  &\ding{51} & \ding{51} & \ding{51} & 54.76 (+0.99)\\
        \multirow{7}{*}{} & 8 & \ding{51} &\ding{51} & \ding{51} & \ding{51} & \textbf{55.20 (+1.43)} \\
        \hline
        \multirow{7}{*}{InstructBLIP-FlanT5XL} & 9 &  &  &  &  & 57.60 \\
        \multirow{7}{*}{} & 10 &  &  &  & \ding{51} & 58.18 (+0.58) \\
        \multirow{7}{*}{} & 11 & \ding{51} &  & & \ding{51} & 58.49 (+0.89) \\
        \multirow{7}{*}{} & 12 & \ding{51} & \ding{51} & & \ding{51} & 59.26 (+1.66)\\
        \multirow{7}{*}{} & 13 & \ding{51} & & \ding{51} & \ding{51} & 58.86 (+1.26) \\
        \multirow{7}{*}{} & 14 & \ding{51} &\ding{51} & \ding{51} &  & 58.95 (+1.35) \\
        \multirow{7}{*}{} & 15 & &\ding{51} & \ding{51} & \ding{51} & 59.07 (+1.47)\\
        \multirow{7}{*}{} & 16 & \ding{51} &\ding{51} & \ding{51} & \ding{51} & \textbf{59.49 (+1.89)} \\
        \hline
    \end{tabular}
    \caption{The results of ablation studies on the design of our proposed framework. The GS knowledge source is adopted here. The ablation experiments are conducted based on RACC-\texttt{homo} with the above two types of MLLMs. }
    \label{tab: ablation studies}
\end{table*}

\section{Details of Different Knowledge Sources}
In this section, we further introduce more details of the two knowledge sources, \emph{i.e.} GS and WIT. 

First of all, the word count in knowledge source documents directly affects the efficiency during training and inference in RAG applications. Therefore, we first present some statistics of the document word counts for these two knowledge sources in Table \ref{tab: word counts}.

We employ FLMR and PREFLMR to retrieve documents from these knowledge sources. FLMR has been trained on OK-VQA and GS, thus it has a strong capability to retrieve relevant documents from the GS knowledge source. PREFLMR is an upgraded version of FLMR, equipped with powerful capabilities for retrieving multimodal documents that FLMR lacks. The detailed retrieval results for OK-VQA with different knowledge sources are shown in Table \ref{tab: prrecalls}. 

Firstly, from Table \ref{tab: prrecalls}, we can tell that there is obviously a number of irrelevant documents among the retrieved documents in different knowledge sources. 

Second, Table \ref{tab: k and prrecall} shows that the quality of the GS knowledge base is superior to that of the WIT knowledge base. The PPRecall@1 metric for the GS knowledge base reaches 63.59\%, indicating that the document with the highest confidence retrieved has a high probability of containing the answer to the question. The PRRecall@5 metric even reaches 88.32\%. These data highlight the advantages of GS as a knowledge source specifically designed for OK-VQA.

\begin{table}[h]
    \centering
    \begin{tabular}{c | c c}
        \hline
        {Knowledge Source} & Metric & Statistic \\
        \hline
        \multirow{7}{*}{GS} & Mean & 59.76 \\
        \multirow{7}{*}{} & Std & 46.05 \\
        \multirow{7}{*}{} & Min & 10 \\
        \multirow{7}{*}{} & 25\% & 29.0 \\
        \multirow{7}{*}{} & 50\% & 45.0 \\
        \multirow{7}{*}{} & 75\% & 75.0 \\
        \multirow{7}{*}{} & Max & 521 \\
        \hline
        \multirow{7}{*}{WIT} & Mean & 155.71 \\
        \multirow{7}{*}{} & Std & 78.86 \\
        \multirow{7}{*}{} & Min & 24 \\
        \multirow{7}{*}{} & 25\% & 89 \\
        \multirow{7}{*}{} & 50\% & 149 \\
        \multirow{7}{*}{} & 75\% & 214 \\
        \multirow{7}{*}{} & Max & 1122 \\
        \hline
    \end{tabular}
    \caption{The statistical data on document word counts of different knowledge sources. The "25\%" column indicates that 25\% of the documents have a word count below this value, and similarly for other percentiles.}
    \label{tab: word counts}
\end{table}

\begin{table}[h]
    \centering
    \begin{tabular}{c | c | c | c c}
        \hline
         \multirow{2}{*}{Knowledge Source} &  \multirow{2}{*}{Retriever} &  \multirow{2}{*}{K} & \multicolumn{2}{c}{PRRecall@K (\%)} \\
        & & & Train & Val \\
        \hline
        \multirow{6}{*}{GS} & \multirow{6}{*}{FLMR} & 1 & 71.56 & 63.59 \\
        \multirow{6}{*}{} & \multirow{6}{*}{}& 2 & 83.77 & 77.07\\
        \multirow{6}{*}{} & \multirow{6}{*}{}& 3 & 88.63 & 82.85\\
        \multirow{6}{*}{} & \multirow{6}{*}{}& 4 & 91.03 & 86.41\\
        \multirow{6}{*}{} & \multirow{6}{*}{}& 5 & 92.82 & 88.32\\
        \multirow{6}{*}{} & \multirow{6}{*}{}& 10 & 95.99 & 93.63\\
        \hline
        \multirow{6}{*}{WIT} & \multirow{6}{*}{PREFLER} & 1 & 37.09 & 35.85 \\
        \multirow{6}{*}{} & \multirow{6}{*}{}& 2 & 48.26 & 46.76 \\
        \multirow{6}{*}{} & \multirow{6}{*}{}& 3 & 54.89 & 53.27 \\
        \multirow{6}{*}{} & \multirow{6}{*}{}& 4 & 59.37 & 57.67\\
        \multirow{6}{*}{} & \multirow{6}{*}{}& 5 & 62.89 & 60.93\\
        \multirow{6}{*}{} & \multirow{6}{*}{}& 10 & 72.16 & 70.41\\
        \hline
    \end{tabular}
    \caption{The detailed retrieval results of different knowledge sources. PRRecall@K measures whether the retrieved $K$ documents contain at least one pseudo-relevant document..}
    \label{tab: prrecalls}
\end{table}

\section{Investigation of hyperparameter $K$}

The hyperparameter $K$ represents the number of retrieved documents used for each image-question pair. The value of $K$ is typically set to 5 in this paper, consistent with RAVQA-v2. We also investigated the impact of different values of $K$ across various knowledge sources and the results are shown in Table \ref{tab: k and prrecall}. 

On the GS knowledge base, when $K = 1$, RACC-\texttt{homo} achieved a performance of 59.65\%, surpassing the 59.49\% performance when $K = 5$. This may be due to the high quality of the GS knowledge base, where the first retrieved document is often likely to contain the information needed to answer the question. In such cases, setting $K$ to a larger value might directly introduce documents with weaker relevance, thereby affecting the performance of the whole framework.

On the WIT knowledge source, we conduct experiments with and without images. In both settings, a larger value of $K$ results in better performance. This may be because there are fewer documents related to the question in the WIT knowledge source, so a larger $K$ allows RACC to extract useful information from more documents. Furthermore, for the same value of $K$, using images results in better performance in both settings, indicating that RACC can obtain valuable information from images. Compared to the results in Table \ref{tab: ravqa reproduction}, we can conclude that RACC has an advantage over RAVQA2 when utilizing multimodal document knowledge sources.

Finally, we can conclude from Table \ref{tab: k and prrecall} that the RACC framework achieves excellent performance across various knowledge sources and different settings of $K$.

\begin{table}[h]
    \centering
    \begin{tabular}{c c c c}
        \hline
        {knowledge source} & {K} & PRRecall@K (\%) & {VQA Accuracy (\%)} \\  
        \hline
        \multirow{3}{*}{WIT} & 1 & 35.85 & 59.11 \\
        \multirow{3}{*}{} & 3 & 53.27 & 59.16 \\
        \multirow{3}{*}{} & 5 & 60.93 & 59.17  \\
        \hline
        \multirow{3}{*}{WIT (Text Only)} & 1 & 35.85 & 58.97  \\
        \multirow{3}{*}{} & 3 & 53.27 & 59.01 \\
        \multirow{3}{*}{} & 5 & 60.93 & 59.04  \\
        \hline
        \multirow{3}{*}{GS} & 1 & 63.59 & 59.65 \\
        \multirow{3}{*}{} & 3 & 82.85 & 59.35 \\
        \multirow{3}{*}{} & 5 & 88.32 & 59.49 \\
        \hline
    \end{tabular}
    \caption{RACC-\texttt{homo}'s experimental results with varying hyperparameter $K$ across different knowledge sources. The MLLM used here is InstructBLIP-FlanT5XL. }
    \label{tab: k and prrecall}
\end{table}

\section{Pseudo Code of RGCA method}
Our proposed RGCA method is applied during the forward computation process of the cross-attention mechanism, aiming to ensure that compressed prompts corresponding to documents with higher retrieval scores receive more attention in the retrieval-guided cross-attention blocks. The pseudo code for the forward function of the retrieval-guided cross-attention block is as follows:

\begin{listing}[hb]
\caption{Pseudo code of RGCA}
\label{lst:listing}%
\begin{lstlisting}[language=Python]
def forward(self, x, context, r_scores):
    # projecting inputs into q, k, v
    # transform q, k, v into multi-head forms
    # expanding the retrieval scores to a certain shape
    r_scores = repeat(r_scores, 'b x y -> (b h) (x m) (y n)', h=self.heads, m=x.shape[1], n=context.shape[1])

    # calculating the similarities between queries and keys
    sim = einsum('b i d, b j d -> b i j', q, k) * self.scale
    sim = sim * r_scores
    attn = sim.softmax(dim=-1)

    # calculating the outputs
\end{lstlisting}
\end{listing}

\end{document}